\pdfoutput=1

\documentclass[11pt]{article}

\usepackage{emnlp2021}

\usepackage{times}
\usepackage{latexsym}
\usepackage{ulem}
\usepackage{diagbox}

\usepackage[T1]{fontenc}

\usepackage[utf8]{inputenc}

\usepackage{microtype}

\usepackage{hyperref}
\usepackage{booktabs,siunitx}
\usepackage{CJKutf8}
\usepackage[nointegrals]{wasysym}
\usepackage{adjustbox}
\usepackage{multirow}
\usepackage{xcolor}
\usepackage{soul}
\usepackage{tabularx,ragged2e}
\usepackage{arydshln}
\usepackage{soul}
\usepackage{xcolor}
\usepackage{amsmath}
\usepackage{mathtools,amssymb,lipsum}
\newcolumntype{C}{>{\Centering\arraybackslash}X} 

\title{BERT-$\beta$: A Proactive Probabilistic Approach to Text Moderation}

\author{Fei Tan,
Yifan Hu,
Kevin Yen,
Changwei Hu \\
Yahoo Research, New York, NY, USA \\
\{fei.tan, yifanhu, kevinyen, changweih\}@yahooinc.com
} 

\begin{document}
\maketitle
\begin{abstract}
Text moderation for user generated content, which helps to promote healthy interaction among users, has been widely studied 
and many machine learning models have been proposed. In this work, we explore an alternative perspective
by augmenting reactive reviews with proactive forecasting. 
Specifically, we propose a new concept {\it text toxicity propensity} to characterize the extent to which a text tends to attract toxic comments. Beta regression is then introduced to 
do the probabilistic modeling, 
which is demonstrated to function well in comprehensive experiments. We also propose an explanation method to communicate the model decision clearly. 
Both propensity scoring and interpretation benefit text moderation in a novel manner. 
Finally, the proposed scaling mechanism for the linear model offers useful insights beyond this work.
\end{abstract}

\section{Introduction}
Text moderation is essential for
maintaining a non-toxic online community 
for media platforms \cite{nobata2016abusive}. 
Many efforts from both academia and industry
have been made to address this critical problem.
Recently, the most prototypical thread is to 
do sophisticated feature engineering or
develop powerful learning algorithms \cite{nobata2016abusive,badjatiya2017deep,bodapati2019neural,tan2020tnt,tran2020habertor}.
Automatic comment moderation schemes 
plus human review
are certainly the cornerstone of the fight against toxicity.

These existing works, however, are {\it reactive} approaches to 
handling user generated text in response to the publication of new articles.
In this paper, we revisit this challenge from a {\it proactive} perspective.
Specifically, 
we introduce a novel concept {\it text toxicity propensity}
to quantify how likely an article is prone to incur toxic comments.
This is a proactive outlook index for news articles prior to the publication, 
which differs radically from the existing reactive approaches to comments.

In this context, reactive describes comment-level moderation algorithms after the publication of news articles (e.g., Perspective \cite{perspective}), which quantifies whether comments are toxic and should be taken down or sent for human review. Proactive emphasizes article-level moderation effort before the publication (without access to comments), which forecasts how likely articles are to attract toxic comments in the future and gives suggestions (e.g., rephrase news articles properly) in advance. Our work can be viewed as the first machine learning effort for a proactive stance against toxicity. 



Formally, we propose a probabilistic approach based on 
Beta distribution \cite{beta_distribution}
to regress article toxicity propensity on article text.
For previously published news articles with comments, 
we take the average of comments' toxicity scores 
as the ground-truth label 
for model learning. The effectiveness of this approach is shown in 
both test set and human labeling.
We also develop a scheme that can provide convincing explanation to the decision of the deep learning model.

\section{Related Works}


To our best knowledge, there's no prior
research thread on proactive text moderation.
Nonetheless, many reactive approaches have been 
explored including hand-crafted feature engineering \cite{chen2012detecting,warner2012detecting,nobata2016abusive}, neural networks \cite{badjatiya2017deep,pavlopoulos2017deeper,agrawal2018deep,zhang2018detecting} and 
transformer variants
\cite{bodapati2019neural,tan2020tnt}.


Recently, context, in the form of parent posts, has been studied but it is 
only viewed as regular text snippets for lifting the performance of toxicity classifiers \cite{pavlopoulos2020toxicity} while screening posts. Our work instead focuses on predicting the proactive toxicity propensity of articles {\em before} they receive user comments.

Beta distribution is usually utilized as a priori in Bayesian statistics. The most popular example in natural language processing is Topic Model \cite{blei2003latent}, where the multivariate version of beta distribution (a.k.a. Dirichlet distribution) generates parameters of mixture models. 
Beta regression is originally proposed for modeling rate and 
proportion data \cite{ferrari2004beta} by parameterizing mean and 
dispersion and regressing parameters of interest.
It has been applied to evaluate grid search parameters in optimization \cite{mckinney2019classification}, model emotional dimensions \cite{aggarwal2020exploration} and statistical processes of child-adult linguistic coordination and alignment \cite{misiek2020development}.


\section{Beta Regression\label{sec_beta}}
In this work, 
both comment toxicity score and the derived article toxicity propensity score (to be detailed in the subsequent section \ref{sec_dataset}) 
range from $0$ to $1$. Empirically, 
their distributions exhibit an asymmetry and may not be modelled well with the Gaussian distribution (Figs. \ref{train_dist} and \ref{score_articles} of Appendix \ref{sec:appendix}). 
Furthermore, comment toxicity score distributions of 
individual articles vary with article content as shown in Fig. \ref{score_articles} of Appendix \ref{sec:appendix}. 
Modelling the entire distribution of an article comment
toxicity scores is thus a reasonable approach. Beta distribution is very flexible and it can model quite 
a wide range of well-known distribution families from symmetric 
uniform ($\alpha=\beta=1$) and bell-shaped distributions 
($\alpha=\beta=2$) to asymmetric shapes ($\alpha\neq\beta$).

In this context, the toxicity propensity score $y$ is assumed to follow the Beta distribution with 
probability density function (pdf):
\begin{equation}
\begin{split}
p(y|\alpha, \beta) & = \mathrm{Beta}(\alpha, \beta) 
 = \frac{y^{\alpha - 1}(1-y)^{\beta - 1}}{B(\alpha, \beta)} \\
\end{split}
\label{original_beta}
\end{equation}
where $\alpha$ and $\beta$ are two positive shape parameters to control the distribution.
$B(\alpha, \beta)$ is the normalization constant and support $y$ meets $y \in [0,1]$.
Eq. \ref{original_beta} holds the probabilistic randomness given $\alpha$ and $\beta$, 
we thus impose a regression structure of 
them on text content. 

Formally, given a training set $\mathcal{D} = \{(\mathbf{x_n}, y_n)\}_{n=1}^N$ with raw text feature vector $\mathbf{x_n}$ and label $y_n$ for sample $n$, we apply feature engineering or text embedding $g(\cdot)$ and then regress $\alpha_n (>0)$ and $\beta_n (>0)$ on $g(\mathbf{x_n})$ respectively as 
\begin{equation}
\begin{split}
   \mathrm{log}(\alpha_n) = f_{\alpha}(g(\boldsymbol{x}_n)) \\
   \mathrm{log}(\beta_n) = f_{\beta}(g(\boldsymbol{x}_n))
\end{split}
\label{alpha_beta}
\end{equation}
where $f_{\alpha}(\cdot)$ and $f_{\beta}(\cdot)$ are learned jointly.
$g(\cdot)$ can be either pre-fixed or learned together with $f_{\alpha}(\cdot)$ and $f_{\beta}(\cdot)$, which is detailed in the subsequent section. Specifically, the learning procedure of $f_{\alpha}(\cdot)$, $f_{\beta}(\cdot)$ and $g(\cdot)$ (if applicable) is to minimize loss
\begin{equation}
\begin{split}
    \mathcal{L} & = -\frac{1}{N}\sum_{n=1}^N\mathrm{log}\big(p(y_n|\alpha_n, \beta_n)\big) \\
\end{split}
\label{loss}
\end{equation}
Substituting Eqs. \ref{original_beta} and \ref{alpha_beta} into it gives the final objective function. 

In the inference phase,
with learned $f_{\alpha}(\cdot)$, $f_{\beta}(\cdot)$ and $g(\cdot)$, 
$\alpha_m$ and $\beta_m$ for a new sample $\mathbf{x_m}$ can be readily derived from Eq. \ref{alpha_beta}. 
We 
take the mean of Eq. \ref{original_beta}  as a point estimator: $\widehat{y_m} = \frac{\alpha_m}{\alpha_m + \beta_m}$ because 
we are predicting the average toxicity.

\section{Experiments}

\subsection{Dataset}\label{sec_dataset}
We collect a dataset of articles published on
Yahoo media outlets, which are all written in English. 
We also exclude articles with low comment volume to make the distribution learning reliable. The number of comments for 99\% of the analyzed articles lie in [10, 8K], with 25\% quantile of 20, median of 50 and mean of 448.
The employed dataset is then
split into training, validation and test parts based on
the publishing date with ratio of 8:1:1 as
described in Table \ref{Dataset}.
It's worthwhile to note that input text $\mathbf{x_n}$ is the concatenation of article title and text body.
The toxicity propensity score $y_n$ of article $n$ is defined as
the average toxicity score of all associated comments.
Comments are scored by Google's Perspective \cite{perspective}, which lies in [0, 1].
Perspective intakes user generated text and outputs toxicity probability. It’s a convolutional neural net \cite{noever2018machine} trained on a comments dataset \footnote{https://www.kaggle.com/c/jigsaw-toxic-comment-classification-challenge} of wikipedia labeled by multiple people per majority rule.

\begin{table}[h!]
\setlength{\tabcolsep}{2pt}
\centering
\caption{Basic statistics of dataset breakdown}
\begin{adjustbox}{width=1\columnwidth}
\tiny
\begin{tabular}{c|c|c|c}
    & Training &  Validation & Test  \\
  \hline
   Sample Size & 536,711  & 70,946 & 70,946   \\
   Publishing Date & 2004 - 05/2020   & 05/2020-06/2020  & 06/2020-09/2020     \\
\end{tabular}
\end{adjustbox}
\label{Dataset}
\end{table}

\subsection{Experiment Setup}
In Eq. \ref{alpha_beta}, we set both 
$f_{\alpha}(\cdot)$ and $f_{\beta}(\cdot)$ to 
single-layer neural networks. 
For $g(\cdot)$,
we experiment with 
either Bag of Words (BOW) or BERT embedding (BERT) \cite{devlin2019bert}.
Specifically, we take uni-gram and bi-gram words sequence 
and compute the corresponding Term Frequency-Inverse  Document  Frequency  (TF-IDF) vectors, 
which leads to around $5.8$ million tokens for BOW. For BERT, we take the 
base version and then fine-tune $f_{\alpha}(\cdot)$ and $f_{\beta}(\cdot)$ on top of the [CLS] embedding, which ends up with $110$ million parameters.
If input text exceeds the maximum length ($510$ as [CLS] and [SEP] are reserved), we adopt a simple yet effective truncation scheme \cite{sun2019fine}. Specifically, we empirically select the first 128 and the last 382 tokens for long text. The rationale is that the informative snippets are more likely to reside in the beginning and end. Batch size is 16 and learning rate is $1e-5$ for Adam optimizer \cite{kingma2015adam}. They are called BOW-$\beta$ and BERT-$\beta$ for short.

\subsection{Baseline Methods and Metrics}

We compare with the linear regression method using BOW features, as well as the BERT base model. Both are combined with one of two loss functions, Mean Absolute Error 
(MAE) or Mean Squared Error (MSE). We call them BOW-MAE, BOW-MSE, BERT-MAE, BERT-MSE, respectively.
The experiment settings are same as the Beta regression.

Since we are interested in identifying articles of high toxicity propensity, we want to make sure that an article with high average toxicity 
is ranked higher than one with low propensity. Thus in addition to mean absolute error, root mean squared error 
(RMSE) and AUC@Precision-Recall curves (AUC@PR), we measure performance using two ranking metrics, 
Kendall coefficient (Kendall) and Spearman's coefficient (Spearman).

\subsection{Results}
We perform evaluation on the whole test set 
and on human labels.

\subsubsection{Test Set}

Table \ref{performance} details the performance comparisons.
Overall, Beta regression
stands out across different metrics 
regardless of feature engineering due 
to its modeling flexibility.
BERT-based methods also outperform
BOW ones in terms of feature engineering and representation. This is reasonable as the former has 20 times as large parameters as the latter and offers the contextual embedding. 
Interestingly, MAE and MSE schemes don't achieve 
the minimum MAE and RMSE although they 
are dedicated to this goal, 
which might result from the limitation of 
point estimator.

\begin{table}[h!]
\setlength{\tabcolsep}{3pt}
\centering
\caption{Performance comparisons on test set}
\begin{adjustbox}{width=1\columnwidth}
\large
\begin{tabular}{c|cc|cc|cc|cc}
\multirow{2}{*}{}    & \multicolumn{2}{c|}{Kendall} &  \multicolumn{2}{c|}{Spearman}   & \multicolumn{2}{c|}{MAE} & \multicolumn{2}{c}{RMSE} \\
     &  val & test  &  val & test  &  val & test  &  val &  test \\
     
 \hline
 BOW-MAE &  0.332  &  0.314  &  0.488   &  0.464   &  0.076  &  0.081  &  0.095  & 0.100   \\
 BOW-MSE &  0.428  &  0.402  & 0.606   & 0.574   & 0.057 &  0.063   &  0.076  & 0.084    \\
 BOW-$\beta$ & 0.437 & 0.413   &  0.617  & 0.589   & \textbf{0.056}   & 
 \textbf{0.061}  & \textbf{0.075}  & \textbf{0.081}   \\
 \hline
 BERT-MAE & 0.360  & 0.333  & 0.525  &  0.489 & 0.072  & 0.076  & 0.092  &  0.095  \\
 BERT-MSE & 0.442  &  0.423 &  0.621 &  0.598 &  0.070 &  0.073  &  0.089 &  0.093  \\
 BERT-$\beta$ &  \textbf{0.462} & \textbf{0.440}  & \textbf{0.642}  &  \textbf{0.617} & \textbf{0.056}  & 0.065  & \textbf{0.075}   & 0.085   \\

\end{tabular}
\end{adjustbox}
\label{performance}
\end{table}

\subsubsection{Human Labels}
As labels are derived from machine, 
we want a sanity check to ensure that the model decision conforms to human intuition. Namely, when the model classifies an article as having high toxicity propensity, we want to make sure that it correlates well with human judgement. 
To this end, we divide test set into
10 equal buckets with an interval of $0.1$ and 
merge the last 4 buckets into [0.6, 1] due to much fewer articles with score being above $0.6$ (as shown in Fig. \ref{train_dist}). There are thus a total of $7$ buckets [0, 0.1), [0.1, 0.2), 
[0.2, 0.3), [0.3, 0.4), [0.4, 0.5), [0.5, 0.6) and [0.6, 1.0]. We then randomly take $100$ samples per bucket and 
set aside $10\%$ for human training and the remaining
are labelled by the human judges
as the benchmark set.
We recruit two groups of people for independent annotation, 
which are required to pick one  
from five levels (a reasonable balance between smoothness and accuracy for manually labeling toxicity propensity per judges' suggestion) to describe the propensity extent to which an article is likely to attract toxic comments: 
Very Unlikely (VU), Unlikely (U), Neutral (N), Likely (L) and Very Likely (VL).
Table \ref{human_review_matrix} is the confusion matrix showing 
how much two groups of human judges agree with each other. 

\begin{table}[h!]
\centering
\caption{Confusion matrix of two groups of human annotation G1 and G2}
\begin{adjustbox}{width=1\columnwidth}
\tiny
\begin{tabular}{|c|c|c|c|c|c|c|}
\hline
  \diagbox{G1}{G2} &   VU &  U & N &  L & VL & Total \\
  \hline
   VU &   89  &28 &  0  & 23 & 1 & 141\\
   U &  30  & 26 &  0  & 37 & 3  &  96 \\ \cdashline{2-3}
   N & 0   & 1 &  0  & 3 & 1 &  5 \\ \cdashline{5-6}
   L &  31  & 25 &  0  & 34 & 56  & 146 \\ 
   VL &  18  & 21 & 0   & 87 & 124  & 250 \\
   \hline
   Total &  168  & 101 &  0  & 184  &  185 &  638 \\
   \hline
\end{tabular}
\end{adjustbox}
\label{human_review_matrix}
\end{table}

\begin{figure*}[h]
\centering
\includegraphics[scale = 0.33]{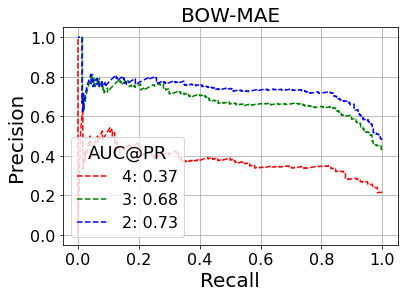}
\includegraphics[scale = 0.33]{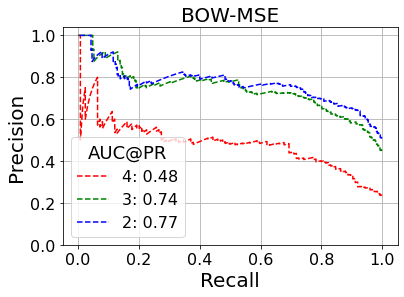}
\includegraphics[scale = 0.33]{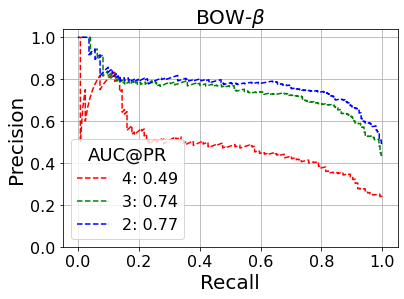}
\includegraphics[scale = 0.33]{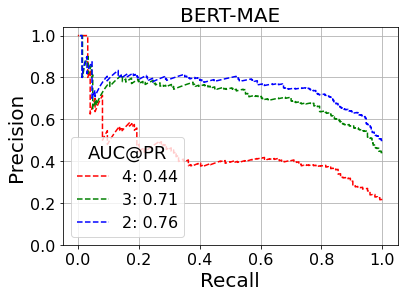}
\includegraphics[scale = 0.33]{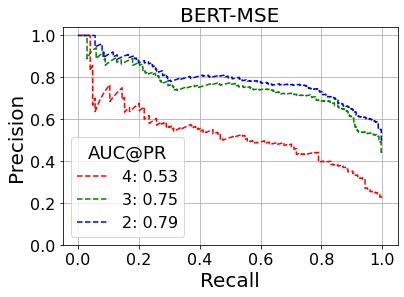}
\includegraphics[scale = 0.33]{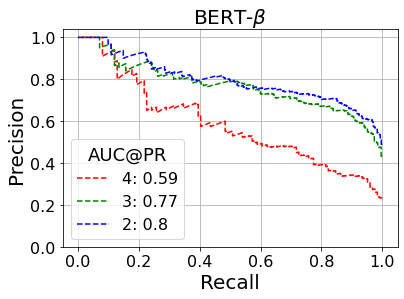}
\caption{PR curves on human labelled data.}
\label{human_pr}
\end{figure*}

Interestingly, even humans don't agree with each other on all examples. Roughly, the perception of the two groups is consist on $74\%$ samples (top left and bottom right boxes in Table \ref{human_review_matrix}). 
Moreover, Cohen's Kappa is about $0.23$ by taking expected chance agreement into account \footnote{\url{https://en.wikipedia.org/wiki/Cohen\%27s_kappa}}. 
In light of this, we jointly score the set by 
assigning $-2$, $-1$, $0$, $1$ and $2$ to VU, U, N, L and VL, respectively. Since each article has two labels, the addition gives an integer score interval [-4,4].
Table \ref{human_perf} reports the performance with human labels as the ground truth, which 
confirms the previous findings that BERT-$\beta$ performs the best.  
Additionally, we pick scores $2$, $3$ and $4$ as thresholds
to monitor precision and recall curves 
(Fig. \ref{human_pr}).
Likewise, the proposed schemes achieve compelling performance widely. 

Taken together, our probabilistic methods agree more with both 
machine and human judgements.

\begin{table}[h!]
\centering
\caption{Performance on human labelled set}
\begin{adjustbox}{width=1\columnwidth}
\tiny
\begin{tabular}{c|c|c|c}
    & BOW-MAE &  BOW-MSE &  BOW-$\beta$  \\
  \hline
   Kendall &  0.402   & 0.481 &  0.491    \\
   Spearman  & 0.562  & 0.635 & 0.649 \\
   \hline
    & BERT-MAE & BERT-MSE & BERT-$\beta$ \\
  \hline
  Kendall  & 0.441 &  0.508  & \textbf{0.522} \\
  Spearman  & 0.599 & 0.665 & \textbf{0.679} 
\end{tabular}
\end{adjustbox}
\label{human_perf}
\end{table}


\subsection{Explanation}
\label{Explanation}
As we focus on the pre-publication
text moderation, a reasonable explanation is 
an essential step to convince stake-holders of 
subsequent operations. 
For BERT-$\beta$ explanation, 
we adopt gradient-based 
saliency map variants from computer vision
\cite{simonyan2013deep, shrikumar2017learning}. 
We compute the gradient $\nabla f(\mathbf{x})$ with respect to input tokens embedding $\mathbf{e(x)}$,
where $f(\mathbf{x})=\alpha(\mathbf{x}) / (\alpha(\mathbf{x}) + \beta(\mathbf{x}))$ is the mean prediction for sample $\mathbf{x}$ (Section~\ref{sec_beta}), 
and 
$\mathbf{x}=(t_1, t_2, \cdots, t_L)$ where $t_l (l=1,2,\cdots, L)$ 
is a single token.
The element of $\nabla f(\mathbf{x})$ is partial derivative $\frac{\partial{f}}{\partial{\mathbf{e}(t_l)}}(\mathbf{x})$ to measure the token-level contribution to the scoring.
The explanation is conducted by assuming
the article is controversial, and we want to figure out which words cause 
some comments to be toxic. So it also makes sense to maximize the maximum toxicity of the comments. 
We thus experiment with 
$f(\mathbf{x})=(\alpha(\mathbf{x}) - 1) / (\alpha(\mathbf{x}) + \beta(\mathbf{x}) - 2)$, 
which is the mode
(corresponding to the peak in the PDF of Beta distribution) under reasonable assumption ($\alpha, \beta > 1$). We denote the resulting scheme by subscript "mode".

For saliency map (SM) \cite{simonyan2013deep}, the metric is
$\Vert\frac{\partial{f}}{\partial{\mathbf{e}(t_l)}}(\mathbf{x})\Vert_2$ without direction. A variant is dot
product (DP) between token embedding and gradient element  $\mathbf{e}(t_l)^T \cdotp \frac{\partial{f}}{\partial{\mathbf{e}(t_l)}}(\mathbf{x})$ with direction 
\cite{shrikumar2017learning}. We also propose a hybrid (HB) scheme to take magnitude of SM and direction of DP to form a new metric. 
We perform an ablation study (AS) to delete single token $t_l$ alternately and 
then compute the score discrepancy between original $\mathbf{x}$ and $\mathbf{x}_{\neg{l}}$ as well.
As a reference, we examine the regression coefficients (RC) of linear BOW-MSE, which are easy to check for explaining the contribution of corresponding words.

A few well-trained human judges are recruited to tag $k$ (example-specific, determined by annotators) most 
important words.
We then prioritize tokens with different metrics
and pick top $k$ ones as candidates. 
Hit rate (proportion of human annotated tokens covered 
by schemes) is used to compare different tools. We 
take $1,000$ examples for human review and compute the average hit rate, 
as compared in Table \ref{explanation_perf}.

All schemes for BERT-$\beta$ are much better than linear scheme RC, 
which is consistent with the predictive performance discrepancy.
SM and HB are close and 
outperform black-box ablation study, 
which implies the valuable role of model-aware gradients in the explanation.
DP is inferior to AS and seems not consistent with human annotation as well as 
other gradient based methods. 
In practice, we take SM for the explanation 
(Appendix \ref{sec:appendix2}) due to its out-performance and simplicity. As expected,
mode (SM$_{\text{mode}}$) covers more annotated words than mean (SM) on average (more discussions in Appendix \ref{sec:appendix3}).

\vspace{-0.2cm}
\begin{table}[h!]
\centering
\caption{Performance (average hit rate) comparison}
\small
\begin{tabular}{ccccc|c}
     SM  & DP & HB &  AS & RC & SM$_{\text{mode}}$  \\
  \hline
    {\bf 0.549}   & 0.430 & 0.543 & 0.467 & 0.382 & \color{teal}\bf{0.553}
\end{tabular}
\label{explanation_perf}
\end{table}
\vspace{-0.2cm}

\section{Additional Study}
Linear regression (BOW-MSE) is inferior to
BERT-$\beta$.
Nonetheless, it is much faster in training, inference and
explanation as it is about 20 times as small as BERT-$\beta$. Thus, we investigate if the performance of the linear model could be improved for
industrial deployment.

Inspired by NBSVM \cite{wang2012baselines},
we scale TF-IDF vectors of BOW-MSE by 
a weight vector defined as
$    \mathbf{w} = \frac{\big(\mathbf{\tau(X)}-\overline{\mathbf{\tau(X)}}\big)^T \big(\mathbf{y}-\overline{\mathbf{y}}\big)}{\big\Vert\mathbf{\tau(X)}-\overline{\mathbf{\tau(X)}}\big\Vert^2}$
where $\mathbf{X}$ is the training corpus. $\mathbf{y} \in [0,1]^{N \times 1}$ 
and $\mathbf{\tau(X)} \in \mathbb{Z}^{N\times M}$ ($M = 5.8$ million) are the training labels and TF-IDF matrix.
$\overline{\mathbf{y}}$ and $\overline{\mathbf{\tau(X)}}$ are their column-wise means.
The pre-computed $\mathbf{w}$ can be viewed as a surrogate of the regression coefficient for 
the linear regression problem, 
which is used to scale TF-IDF of BOW-MSE in both training and inference phases. We call it Naive Bayes Linear Regression (NBLR) for short.

The scaling benefits the performance, as compared in Table \ref{performance_human_improved}. As can be seen, NBLR improves upon BOW-MSE significantly, although it is not as good as BERT-$\beta$.

\begin{table}[h!]
\setlength{\tabcolsep}{2pt}
\centering
\caption{Performance on test set and human labels}
\begin{adjustbox}{width=1\columnwidth}
\large
\begin{tabular}{c|cccc}
\multirow{2}{*}{}    & \multicolumn{2}{c}{Test Set} &  \multicolumn{2}{c}{Human Label}  \\
     &  Kendall & Spearman  &  Kendall & Spearman  \\
 \hline
 BOW-MSE &   0.402 &   0.574 &  0.481   &   0.635     \\
 NBLR & {\bf 0.413} (+.011)  & {\bf 0.588} (+.014)   &   {\bf 0.501} (+.020) &  {\bf 0.656} (+.021)    \\
 BERT-$\beta$  & 0.440  &  0.617  &   0.522  &  0.679      \\
 \hline
 \multirow{2}{*}{}    & \multicolumn{3}{c}{Human Label} &    \\
 &  AUC@PR at 2 & AUC@PR at 3  &  AUC@PR at 4 &  \\
 \hline
  BOW-MSE &  0.77  &   0.74    &   0.48 &        \\
 NBLR &   {\bf 0.78} (+.01)   & {\bf 0.76} (+.02) & {\bf 0.53} (+.05)   &     \\
  BERT-$\beta$  & 0.80 &   0.77 &  0.59   &        \\
 \hline
\end{tabular}
\end{adjustbox}
\label{performance_human_improved}
\end{table}


\section{Discussion and Future Work}
Our work can benefit text moderation.
The proactive propensity offers a toxicity outlook for comments, which could be utilized in multiple ways. 
For example, stricter moderation rules are enforced
for articles that are predicted to have a high toxicity propensity.
Furthermore, the propensity could be used
as an additional feature for 
the downstream reactive toxicity recognition models, as well as for allocation of appropriate human resources. 

The explanation tool can also be used to remind
editors to rephrase some controversial words to 
mitigate the odds of attracting toxic comments.
Text moderation is an important yet challenging task,
our proactive work is attempting to open up a new perspective to augment the traditional reactive procedure.
Our current model, however, is not perfect as shown 
by article b in Fig. \ref{score_articles} of Appendix \ref{sec:appendix} where the learned distribution doesn't fit well the observed histogram.
Technically, NBLR is an encouraging lightweight extension to Linear Regression. 
Likewise, we will continue to work towards the improvement 
of the non-linear Beta regression. 

\section{Conclusion}
We approach text moderation by developing
a well-motivated probabilistic model to learn 
a proactive toxicity propensity.
An explanation scheme is also proposed to 
visually explain the connection between 
this new prospective score, and
text content. Our experiment shows the superior
performance of the proposed BERT-$\beta$ algorithm, 
compared with a number of baselines, 
in predicting both the average toxicity score,
and the human judgement.



\bibliography{cone_short.bib}
\bibliographystyle{acl_natbib}

\clearpage

\appendix

\section{Toxicity Score and Beta Distribution}
\label{sec:appendix}

\begin{figure}[h]
\centering
\includegraphics[scale = 0.5]{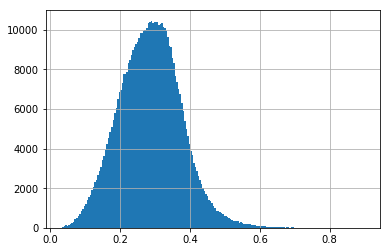}
\caption{Toxicity propensity score (mean comment toxicity scores) distribution of news articles.}
\label{train_dist}
\end{figure}

\begin{figure}[h]
\centering
\includegraphics[scale = 0.5]{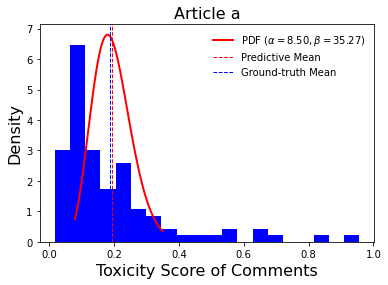}
\includegraphics[scale = 0.5]{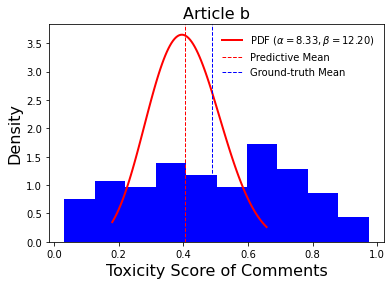}
\caption{Toxicity score histogram density of comments for articles a (top) and b (bottom). Solid red lines represent predictive beta distribution for individual articles.}
\label{score_articles}
\end{figure}

The distribution of news articles' toxicity propensity score is reported in
Fig. \ref{train_dist}. 
Comment score distributions of two articles with predictive distribution are given in Fig. \ref{score_articles}.

\section{SM Explanation Examples}
\label{sec:appendix2}
\begin{figure*}[h]
\centering
\includegraphics[width=\textwidth]{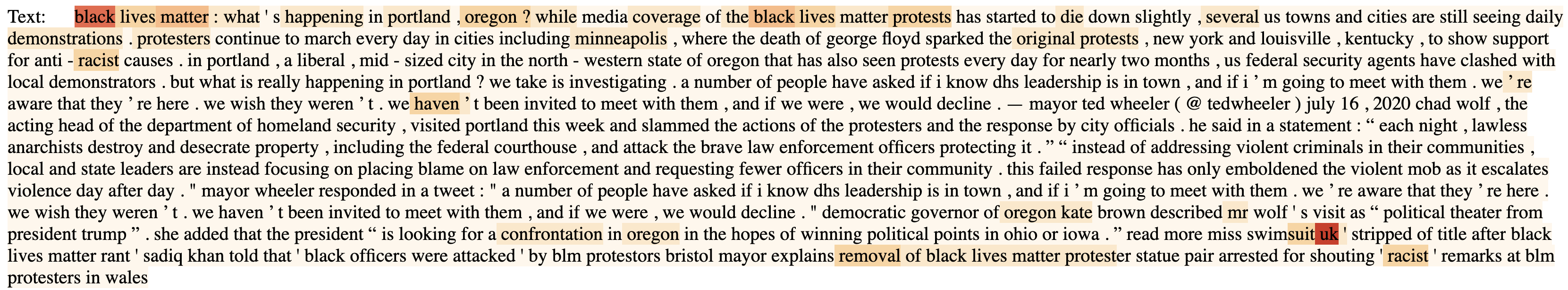}
\includegraphics[width=\textwidth]{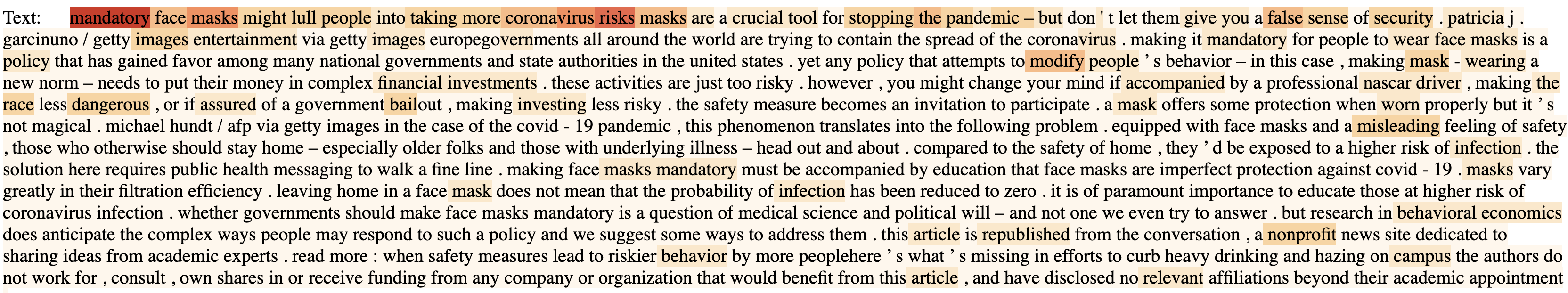}
\caption{Model explanation examples.}
\label{explanation_examples}
\end{figure*}

We pick two samples from the test set and then
leverage SM in section \ref{Explanation} to 
highlight key words for the illustration purpose, as shown in 
Fig. \ref{explanation_examples}.
The color intensity is proportional to the 
normalized saliency map value. 
The darker the color of a token is, 
the more important it's to the scoring. 
There's also a positional bias towards the first sentence as it's the article title.

\section{BERT-$\beta$ mode}
\label{sec:appendix3}
We also explore the mode of BERT-$\beta$ as a point estimator and compare it with the mean. 
Table \ref{mode_performance_human_improved} details the performance discrepancy 
between the test set and human labels. 
For the toxicity propensity prediction in the test set, 
it does make sense for mean to slightly outperform mode as ground-truth labels are the score mean of comments. 
When it comes to human labels and explanation, people
annotate news articles based on the perceived controversial words most likely to
incur toxic comments. Mode is thus able to capture the worst case better and 
agrees more with human annotations. 
This finding is in  line with 
the better explanation performance, as compared in Table \ref{explanation_perf}.

\begin{table*}[h!]
\centering
\caption{Performance of BERT-$\beta$ point estimators on test set and human labels}
\begin{adjustbox}{width=1\textwidth}
\large
\begin{tabular}{c|cccc|cc}
\multirow{2}{*}{}    & \multicolumn{4}{c|}{Test Set} &  \multicolumn{2}{c}{Human Label}  \\
     &  Kendall & Spearman & RMSE & MAE  &  Kendall & Spearman  \\
 \hline
 mean &  \bf{0.440}  &  \bf{0.617}  &  \bf{0.065}   &   \bf{0.085}  & 0.522 & 0.679   \\
 mode &  0.439 &  0.614  &  0.077   &   0.099 & \bf{0.543}  &  \bf{0.704}   \\
 \hline
 \multirow{2}{*}{}    & \multicolumn{3}{c}{Human Label} &    \\
 &  AUC@PR at 2 & AUC@PR at 3  &  AUC@PR at 4 &  \\
 \hline
 mean &  0.8  &    0.77 &   0.59  &        \\
 mode &  \bf{0.82} &  \bf{0.79}  &  \bf{0.63} &     \\
 \hline
\end{tabular}
\end{adjustbox}
\label{mode_performance_human_improved}
\end{table*}





\end{document}